\documentclass[10pt,twocolumn,letterpaper]{article}
\pdfoutput=1
\usepackage{cvpr}
\usepackage{times}
\usepackage{epsfig}
\usepackage{graphicx}
\usepackage{amsmath}
\usepackage{amssymb}
\usepackage{array}
\usepackage{enumitem}
\usepackage[footnotesize]{subfigure}

\usepackage[lined,linesnumbered,ruled]{algorithm2e}

\SetCommentSty{mycommfont}

\usepackage[pagebackref=true,breaklinks=true,letterpaper=true,colorlinks,bookmarks=false]{hyperref}

\cvprfinalcopy 

\def\cvprPaperID{2334} 

\ifcvprfinal\fi
\begin{document}

\title{Visual Translation Embedding Network for Visual Relation Detection}

\author{Hanwang Zhang$^{\dagger}$, Zawlin Kyaw$^\ddagger$, Shih-Fu Chang$^\dagger$, Tat-Seng Chua$^\ddagger$\\
	$^\dagger$Columbia University, $^\ddagger$National University of Singapore\\
	\{hanwangzhang, kzl.zawlin\}@gmail.com; {sfchang@ee.columbia.edu}; {dcscts@nus.edu.sg}\\
}

\maketitle

\begin{abstract}
Visual relations, such as ``person ride bike'' and ``bike next to car'', offer a comprehensive scene understanding of an image, and have already shown their great utility in connecting computer vision and natural language. However, due to the challenging combinatorial complexity of modeling subject-predicate-object relation triplets, very little work has been done to localize and predict visual relations. Inspired by the recent advances in relational representation learning of knowledge bases and convolutional object detection networks, we propose a Visual Translation Embedding network (VTransE) for visual relation detection. VTransE places objects in a low-dimensional relation space where a relation can be modeled as a simple vector translation, i.e., subject + predicate $\approx$ object. We propose a novel feature extraction layer that enables object-relation knowledge transfer in a fully-convolutional fashion that supports training and inference in a single forward/backward pass. To the best of our knowledge, VTransE is the first end-to-end relation detection network. We demonstrate the effectiveness of VTransE over other state-of-the-art methods on two large-scale datasets: Visual Relationship and Visual Genome. Note that even though VTransE is a purely visual model, it is still competitive to the Lu's multi-modal model with language priors~\cite{lu2016visual}.  
\end{abstract}
\vspace{-4mm}
\section{Introduction}
\vspace{-1mm}
We are witnessing the impressive development in connecting computer vision and natural language, from the arguably mature visual detection~\cite{he2015deep,ren2015faster} to the burgeoning visual captioning and question answering~\cite{antol2015vqa,bernardi2016automatic}. However, most existing efforts to the latter vision-language tasks attempt to directly bridge the visual model (\eg, CNN) and the language model (\eg, RNN), but fall short in modeling and understanding the relationships between objects. As a result, poor generalization ability was observed as those models are often optimized on specialized datasets for specific tasks such as image captioning or image QA.~\cite{jabri2016revisiting,vinyals2016show}.

\begin{figure}
	\centering
	\includegraphics[width=.9\linewidth]{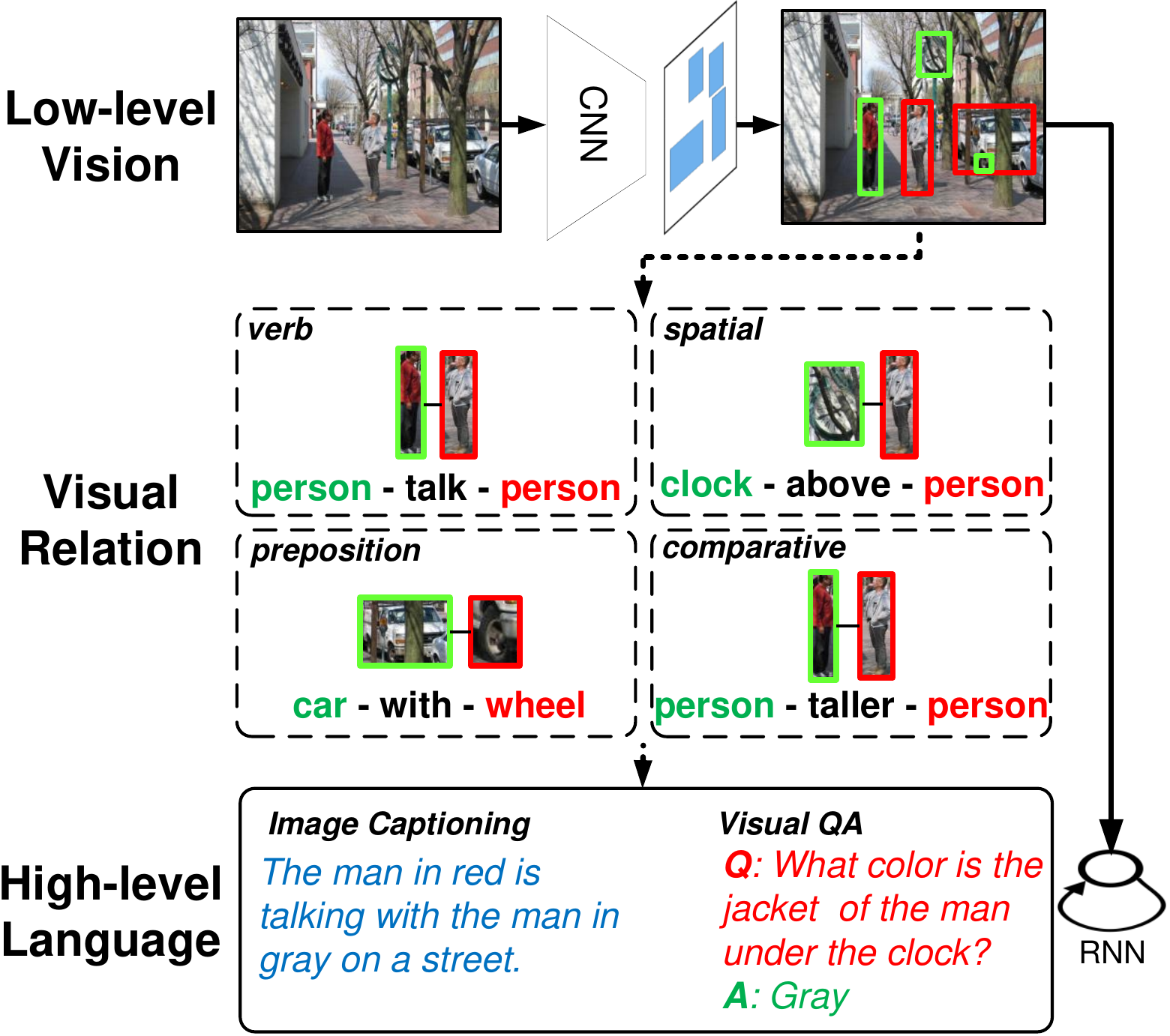}
	\caption{We focus on detecting \emph{visual relations} (dashed boxes in the middle layer) in this paper. Different from the direct connection between low-level vision and high-level language, visual relations offer the direct understanding of object interactions, which provide further semantic information for applications such as image captioning and QA.}
	\label{fig:1}
	\vspace{-2mm}
\end{figure}

As illustrated in Figure~\ref{fig:1}, we take a step forward from the lower-level object detection and a step backward from the higher-level language modeling, focusing on the \textit{visual relations} between objects in an image. We refer to a visual relation as a \texttt{subject-predicate-object} triplet\footnote[1]{When the context is clear, we always refer to object in normal font as a general object and \texttt{object} in teletype to the tail object in a relation.}, where the predicate can be verb (\texttt{person1}-\texttt{talk}-\texttt{person2}), spatial (\texttt{clock}-\texttt{above}-\texttt{person2}), preposition (\texttt{car}-\texttt{with}-\texttt{wheel}), and comparative (\texttt{person1}-\texttt{taller}-\texttt{person2})~\cite{krishna2016visual,lu2016visual}. Visual relations naturally bridge the vision and language by placing objects in a semantic context of what, where, and how objects are connected with each other. For example, if we can detect \texttt{clock}-\texttt{above}-\texttt{person2} and \texttt{person2}-\texttt{wear}-\texttt{jacket} successfully, the reasoning behind the answer ``gray'' to the question asked in Figure~\ref{fig:1} will be explicitly interpretable using dataset-independent inference, \eg, QA over knowledge bases~\cite{dong2015question}, and thus permits better generalization or even zero-shot learning~\cite{krishna2016visual,wang2016fvqa}. 

\begin{figure}
	\centering
	\includegraphics[width=.9\linewidth]{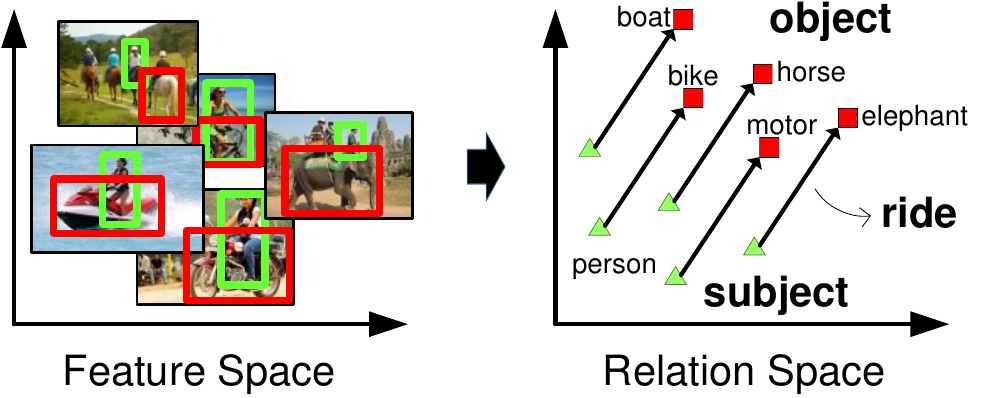}
	\caption{An illustration of translation embedding for learning predicate \texttt{ride}. Instead of modeling from a variety of \texttt{ride} images, VTransE learns consistent translation vector in the relation space regardless of the diverse appearances of subjects (\eg, \texttt{person}) and objects (\eg, \texttt{horse}, \texttt{bike}, \etc.) involved in the predicate relation (\eg, \texttt{ride}).}
	\label{fig:2}
	\vspace{-2mm}
\end{figure}

In this paper, we present a convolutional localization network for visual relation detection dubbed \textbf{V}isual \textbf{Trans}lation \textbf{E}mbedding network (\textbf{VTransE}). It detects objects and predicts their relations simultaneously from an image in an end-to-end fashion. We highlight two key novelties that make VTransE effective and distinguishable from other visual relation models~\cite{lu2016visual,sadeghi2015viske,sadeghi2011recognition}:

\textbf{Translation Embedding}.
Since relations are compositions of objects and predicates, their distribution is much more long-tailed than objects. For $N$ objects and $R$ predicates, one has to address the fundamental challenge of learning $\mathcal{O}(N^2R)$ relations with few examples~\cite{ramanathan2015learning,sadeghi2011recognition}. A common solution is to learn separate models for objects and predicates, reducing the complexity to $\mathcal{O}(N+R)$. However, the drastic appearance change of predicates makes the learning even more challenging. For example, \texttt{ride} appearance largely varies from \texttt{person}-\texttt{ride}-\texttt{bike} to \texttt{person}-\texttt{ride}-\texttt{elephant}. To this end, inspired by Translation Embedding (TransE) in representing large-scale knowledge bases~\cite{bordes2013translating,lin2015learning}, we propose to model visual relations by mapping the features of objects and predicates in a low-dimensional space, where the relation triplet can be interpreted as a vector translation, \eg, \texttt{person}$+$\texttt{ride} $\approx$ \texttt{bike}. As shown in Figure~\ref{fig:2}, by avoiding learning the diverse appearances of \texttt{subject}-\texttt{ride}-\texttt{object} with large variance, we only need to learn the \texttt{ride} translation vector in the relation space, even though the subjects and/or objects can be quite diverse.

\textbf{Knowledge Transfer in Relation}.
Cognitive evidences show that the recognition of objects and their interactions is reciprocal~\cite{chao2000representation,gupta2009observing}. For example, \texttt{person} and \texttt{bike} detections serve as the context for \texttt{ride} prediction, which in turn constrains the articulation of the two objects, and thus benefiting object detection.  Inspired by this, we explicitly incorporate knowledge transfer between objects and predicates in VTransE. Specifically, we propose a novel feature extraction layer that extracts three types of object features used in translation embedding: classeme (\ie, class probabilities), locations (\ie, bounding boxes coordinates and scales), and RoI visual features. In particular, we use the bilinear feature interpolation~\cite{gregor2015draw,jaderberg2015spatial} instead of RoI pooling~\cite{girshick2015fast,ren2015faster} for differentiable coordinates. Thus, the knowledge between object and relation---confidence, location, and scale---can be transfered by a single forward/backward pass in an end-to-end fashion.

We evaluate the proposed VTransE on two recently released relation datasets: Visual Relationship~\cite{lu2016visual} with 5,000 images and 6,672 unique relations, and Visual Genome~\cite{krishna2016visual} with 99,658 images and 19,237 unique relations. We show significant performance improvement over several state-of-the-art visual relation models. In particular, our purely visual VTransE can even outperform the multi-modal method with vision and language priors~\cite{lu2016visual} in detection and retrieval, and a bit shy of it in zero-shot learning.

In summary, our contributions are as follows: 1) We propose a visual relation detection model dubbed Visual Translation Embedding network (VTransE), which is a convolutional network that detects objects and relations simultaneously. To the best of our knowledge, this is the first end-to-end relation detection network; 2) We propose a novel visual relation learning model for VTransE that incorporates translation embedding and knowledge transfer; 3) VTransE outperforms several strong baselines on visual relation detection by a large performance gain.

\begin{figure*}
	\centering
	\includegraphics[width=1\linewidth]{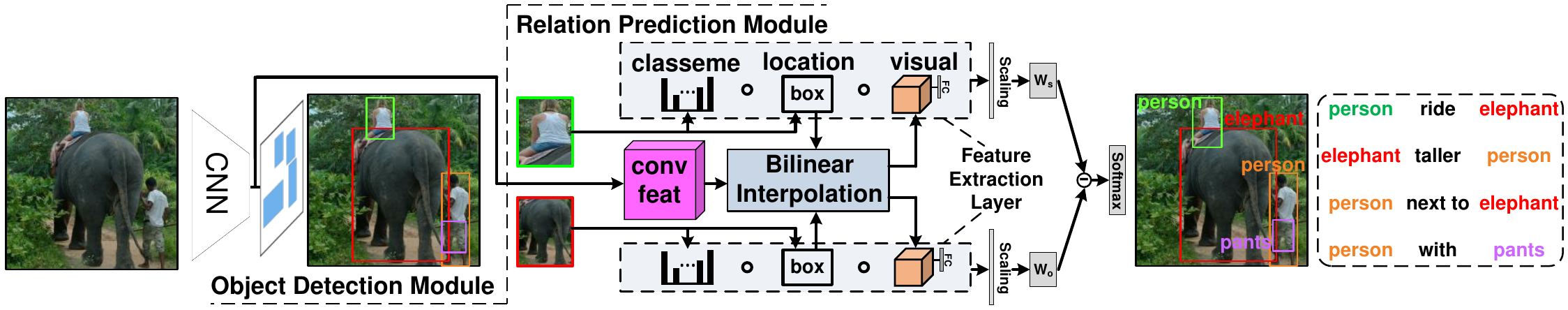}
	\caption{The VTransE network overview. An input image is first through the Object Detection Module, which is a convolutional localization network that outputs a set of detected objects. Then, every pair of objects are fed into the Relation Prediction Module for feature extraction and visual translation embedding. In particular, the visual feature of an object is smoothly extracted from the last convolutional feature map using Bilinear Interpolation. $\circ$ denotes vector concatenation and $\ominus$ denotes element-wise subtraction.}
	\label{fig:3}
	\vspace{-4mm}
\end{figure*}

\section{Related Work}
Our work falls in the recent progress on grounding compositional semantics in an image~\cite{krishna2016visual,plummer2015flickr30k}. It has been shown that high-quality groundings provide more comprehensive scene understanding, which underpins many vision-language tasks such as VQA~\cite{andreas2016deep}, captioning~\cite{karpathy2015deep} and complex query retrieval~\cite{johnson2015image}. Visual relation detection not only ground regions with objects, but also describes their interactions. In particular, our VTransE network draws on recent works in relation learning and object detection. 

\textbf{Visual Relation Detection}.
Different from considering relations as hidden variables~\cite{yao2010modeling}, we relate to explicit relation models which can be divided into two categories: joint model and separate model. For joint models, a relation triplet is considered as a unique class~\cite{atzmon2016learning,farhadi2010every,ramanathan2015learning,sadeghi2011recognition}. However, the long-tailed distribution is an inherent defect for scalability. Therefore, we follow the separate model that learns subject, object, and predicate individually~\cite{desai2011discriminative,gupta2008beyond,sadeghi2015viske,lu2016visual}. But, modeling the large visual variance of predicates is challenging. Inspired by TransE that has been successfully used in relation learning in large-scale knowledge base~\cite{bordes2013translating,lin2015learning}, our VTransE extends TransE for modeling visual relations by mapping subjects and objects into a low-dimensional relation space with less variance, and modeling the predicate as a translation vector between the subject and object. Note that there are works~\cite{atzmon2016learning,lu2016visual,ramanathan2015learning} that exploit language priors to boost relation detection, but we are only interested in visual models.

\textbf{Object Detection}.
VTransE is based on an object detection module composed of a region proposal network (RPN) and a classification layer. In particular, we use Faster-RCNN~\cite{ren2015faster}, which is evolved from its predecessors~\cite{girshick2015fast,girshick2014rich} that requires additional input of region proposals. Note that VTransE cannot be simply considered as appending a relation prediction layer to Faster-RCNN. In fact, we propose a novel feature extraction layer that allows knowledge transfer between objects and relations. The layer exploits the bilinear interpolation~\cite{gregor2015draw,jaderberg2015spatial,johnson2015densecap} instead of the non-smooth RoI pooling in Faster-RCNN and thus the reciprocal learning of objects and predicates can be achieved in a single forward/backward pass. Note that VTransE can be married to any object detection network that contains an RPN such as the very recent SSD~\cite{liu2015ssd} and YOLO~\cite{redmon2015you}. 

\section{Our Approach: VTransE Network}
VTransE is an end-to-end architecture that completes object detection and relation prediction simultaneously. As illustrated in Figure~\ref{fig:3}, it builds upon an object detection module (\eg, Faster-RCNN), and then incorporates the proposed feature extraction layer and the translation embedding for relation prediction. 
\subsection{Visual Translation Embedding}\label{sec:feat}
Given any valid relation, Translation Embedding (TransE)~\cite{bordes2013translating} represents \texttt{subject}-\texttt{predicate}-\texttt{object} in low-dimensional vectors \textbf{s}, \textbf{p}, and \textbf{o}, respectively, and the relation is represented as a translation in the embedding space: \textbf{s} + \textbf{p} $\approx$ \textbf{o} when the relation holds, and \textbf{s} + \textbf{p} $\not\approx$ \textbf{o} otherwise. TransE offers a simple yet effective linear model for representing the long-tail relations in large knowledge databases~\cite{nickel2016review}. To transfer TransE in the visual domain, we need to map the features of the detected objects into the relation space, which is consistent to recent works in visual-semantic embeddings~\cite{akata2015label,frome2013devise}. 

Suppose $\mathbf{x}_s,\mathbf{x}_o\in\mathbb{R}^M$ are the $M$-dimensional features of \texttt{subject} and \texttt{object}, respectively. Besides learning a relation translation vector $\mathbf{t}_p\in\mathbb{R}^{r}$ ($r\ll M$) as in TransE\footnote[2]{In experiments, we tested $r\in\{100,200,...,1000\}$ and found that $r = 500$ is a good default.}, VTransE learns two projection matrices $\mathbf{W}_s,\mathbf{W}_o\in\mathbb{R}^{r\times M}$ from the feature space to the relation space. Thus, a visual relation can be represented as:
\begin{equation}\label{eq:1}\footnotesize{
	\mathbf{W}_s\mathbf{x}_s+\mathbf{t}_p\approx \mathbf{W}_o\mathbf{x}_o.}
\end{equation} 
As in TransE, we can use a large-margin metric learning loss function for relations:
\begin{equation}\label{eq:2}\footnotesize{
	\begin{split}
		\mathcal{L} =&\!\!\!\!\! \sum\limits_{(s,p,o)\in\mathcal{R}}\sum\limits_{(s',p,o')\in\mathcal{R}'} [d(\mathbf{W}_s\mathbf{x}_s+\mathbf{t}_p,\mathbf{\mathbf{W}_o\mathbf{x}_o})+1
		\\
		&- d(\mathbf{W}_s\mathbf{x}_s'+\mathbf{t}_p,\mathbf{\mathbf{W}_o\mathbf{x}_o'})]_+
	\end{split}}
\end{equation}
where $d()$ is a proper distance function, $\mathcal{R}$ is the set of valid relations, and $\mathcal{R}'$ is the set of invalid relations. 

However, unlike the relations in a knowledge base that are generally facts, \eg, \texttt{AlanTuring}-\texttt{bornIn}-\texttt{London}, visual relations are volatile to specific visual examples, \eg, the validity of \texttt{car}-\texttt{taller}-\texttt{person} depends on the heights of the specific car and person in an image, resulting in problematic negative sampling if the relation annotation is incomplete. Instead, we propose to use a simple yet efficient softmax for prediction loss that only rewards the deterministically accurate predicates\footnote[3]{In fact, predicate is multi-labeled, \eg, both \texttt{person}-\texttt{on}-\texttt{bike} and \texttt{person}-\texttt{ride}-\texttt{bike} are correct. However, most relations are single-labeled in the datasets, \eg, 58\% in VRD~\cite{lu2016visual} and 67\% in VG~\cite{krishna2016visual}.}, but not the agnostic object compositions of specific examples:
\begin{equation}\label{eq:3}\footnotesize{
	\mathcal{L}_{rel} =  \sum\limits_{(s,p,o)\in\mathcal{R}}-\log \textrm{softmax}\left(\mathbf{t}^T_p(\mathbf{W}_o\mathbf{x}_o-\mathbf{W}_s\mathbf{x}_s)\right),}
\end{equation}
where the softmax is computed over $p$. Although Eq.~\eqref{eq:3} learns a rotational approximation for the translation model in Eq.~\eqref{eq:1}, we can retain the translational property by proper regularizations such as weight decay~\cite{mikolov2013distributed}. In fact, if the annotation for training samples is complete, VTransE works with softmax (Eq.~\eqref{eq:3}) and negative sampling metric learning (Eq.~\eqref{eq:2}) interchangeably.

The final score for relation detection is the sum of object detection score and predicate prediction score in Eq.~\eqref{eq:3}: $S_{s,p,o} = S_{s}+S_{p}+S_o$.
 
\subsection{Feature Extraction}
We propose a Feature Extraction Layer in VTransE to extract $\mathbf{x}_s$ and $\mathbf{x}_o$. There are three types of features that characterize the multiple facets of objects in relations: 
\\
\textbf{Classeme}. It is an $(N+1)$-d vector of object classification probabilities (\ie, $N$ classes and 1 background) from the object detection network. Classeme is widely used as semantic attributes in various vision tasks~\cite{torresani2010efficient}. For example, in relation detection, classeme is a useful prior for rejecting unlikely relations such as \texttt{cat}-\texttt{ride}-\texttt{person}. 
\\
\textbf{Location}. It is a $4$-d vector $(t_x,t_y,t_w,t_h)$, which is the bounding box parameterization in~\cite{girshick2014rich}, where  $(t_x,t_y)$ specifies a scale-invariant translation and $(t_w, t_h)$ specifies the log-space height/width shift relative to its counterpart \texttt{object} or \texttt{subject}. Take \texttt{subject} as an example:
\begin{equation}\label{eq:x}\footnotesize{
t_x = \frac{x-x'}{w'}, t_y = \frac{y-y'}{h'},t_w = \log\frac{w}{w'}, t_h = \log\frac{h}{h'}} 
\end{equation}
where $(x,y,w,h)$ and $(x',y',w',h')$ are the box coordinates of \texttt{subject} and \texttt{object}, respectively. Location feature is not only useful for detecting spatial or preposition relation, but also useful for verbs, \eg, \texttt{subject} is usually above \texttt{object} when the predicate is \texttt{ride}.
\\
\textbf{Visual Feature}. It is a $D$-d vector transformed from a convolutional feature of the shape $X\times Y\times C$. Although it is as the same size as the RoI pooling features used in Faster-RCNN, our features are bilinearly interpolated from the last conv-feature map, so as to achieve end-to-end training that allows knowledge transfer (cf. Section~\ref{sec:arch}). 

The overall feature $\mathbf{x}_s$ or $\mathbf{x}_o$ is a weighted concatenation of the above three features ($M = N+D+5$), where the weights are learnable scaling layers since the feature contribution dynamically varies from relation to relation. As shown in Figure~\ref{fig:3}, the proposed feature extraction layer couples the Object Detection Module and the Relation Prediction Module.

\subsection{Architecture Details}\label{sec:arch}
A training image for VTransE is labeled with a list of \texttt{subject}-\texttt{predicate}-\texttt{object} triplets, where every unique \texttt{subject} or \texttt{object} is annotated with a bounding box. At testing time, VTransE inputs an image and outputs a set of detected objects and the relation prediction scores for every pair of objects. 

\textbf{Object Detection Network}.
VTransE network starts from the Faster-RCNN~\cite{ren2015faster} object detection network with the VGG-16 architecture~\cite{simonyan2014very}. At training time, we sample a mini-batch cotaining 256 region proposal boxes generated by the RPN of Faster-RCNN, each of which is positive if it has an intersection over union (IoU) of at least 0.7 with some ground truth regions and it is negative if the IoU $<$ 0.3. The positive proposals are fed into the classification layer, where each proposal outputs an $(N+1)$ class probabilities and $N$ bounding box estimations. Then, we perform non-maximum suppression (NMS) for every class with the IoU $>$ 0.4, resulting in 15.6 detected objects on average, each of which has only one bounding box. The reasons of performing NMS for object detection are two folds: 1) we need a specific object class for each region to match with the relation ground truth, and 2) we need to down-sample the objects for a reasonable number of candidate relations. At test time, we sample 300 proposal regions generated by RPN with IoU $>$ 0.7. After the classification layer, we perform NMS with IoU $>$ 0.6 on the 300 proposals, resulting in 15--20 detections per image on average.

\textbf{Bilinear Interpolation}.
By removing the final pooling layer of VGG-16, we use the last convolutional feature map $\mathbf{F}$ of the shape $W' \times H'\times C$ (the pink cube in Figure~\ref{fig:3}), where $C =512$ is the number of channels, $W' = \lfloor \frac{W}{16}\rfloor$, and $H' = \lfloor \frac{H}{16}\rfloor$, where $W$ and $H$ are the width and height of the input image. $\mathbf{F}$ encodes the visual appearance of the whole image and is used for extracting visual features for the object detection and relation prediction.

In order to achieve object-relation knowledge transfer, the relation error should be back-propagated to the object detection network and thus refines the objects. However, the widely-used RoI pooling visual feature in Fast/Faster R-CNN is not a smooth function of coordinates since it requires discrete grid split for the proposal region, resulting in zero coordinate gradients back-propagated from the feature extraction layer.

To this end, we replace the RoI pooling layer with bilinear interpolation~\cite{jaderberg2015spatial}. It is a smooth function of two inputs: the feature map $\mathbf{F}$ and an object bounding box projected onto $\mathbf{F}$, and the output is a feature $\mathbf{V}$ of the size $X\times Y\times C$ (the orange cube in Figure~\ref{fig:3}). Each entry value in $\mathbf{V}$ can be efficiently interpolated from $\mathbf{F}$ in a convolutional way:
\begin{equation}\label{eq:4}\footnotesize{
V_{i,j,c} = \sum\limits^{W'}_{i'=1}\sum\limits^{H'}_{j' = 1}F_{i',j',c}k(i'-G_{i,j,1})k(j'-G_{i,j,2}),}
\end{equation}
where $\mathbf{G}\in\mathbb{R}^{X\times Y \times 2}$ records the positions of the $X\times Y$ grid split in the input bounding box and $k(x) = \max(0,1-|x|)$ is the bilinear interpolation kernel. Note that the grid position $\mathbf{G}$ matrix is a linear function of the input box. Therefore, the gradients from $\mathbf{V}$ can be back-propagated to the bounding box coordinates.

\textbf{Optimization}.
We train the VTransE network end-to-end by stochastic gradient descent with momentum~\cite{kingma2014adam}. We follow the ``image-centric'' training strategy~\cite{ren2015faster}, \ie, the mini-batch arises from a single image that contains many object regions and relations. The loss function is a multi-task loss combining the object detection loss $\mathcal{L}_{obj}$ and the relation detection loss $\mathcal{L}_{rel}$ in Eq.~\eqref{eq:3}, allowing reciprocal learning for objects and relations. In particular, we find that a reasonable loss trade-off is $\mathcal{L}_{obj}+0.4\mathcal{L}_{rel}$. Since object detection and relation prediction have different sample sizes, we normalize $\mathcal{L}_{obj}$ and $\mathcal{L}_{rel}$ by the mini-batch size. 

For model initializations, we pre-train Faster-RCNN on the objects in the relation datasets to initialize the object detection network and randomly initialize the VTransE component with Gaussian weights. For end-to-end training, we also replace the RoI pooling layer in the object detection network with bilinear interpolation. For efficiency, we do not fine-tune the VGG-16 CNN. Generally, we need 2 -- 3 epochs for the model to converge. For a single image that has been resized to the longer side of 720 pixels, the training runs in 2.0 fps and the testing runs in 6.7 fps on a Titan X GPU using Caffe and Python. Note that we can always plug-in faster object detection networks such as SSD~\cite{liu2015ssd} and YOLO~\cite{redmon2015you} for more efficient training and testing.   

\begin{figure}
	\centering
	\includegraphics[width=.8\linewidth]{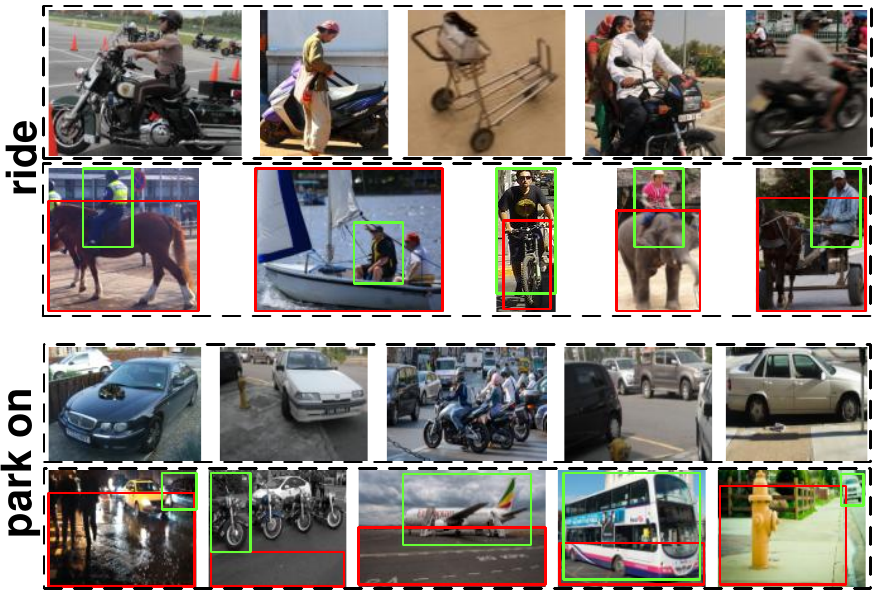}
	\caption{Top 5 confident regions of \texttt{subject} and \texttt{object} retrieved by \texttt{ride} and \texttt{park on} models of JointBox (1st row) and VTransE (2nd row with ground-truth bounding boxes) from VRD.}
	\vspace{-4mm}
	\label{fig:4}
\end{figure}

\section{Experiments}
We are going to validate the effectiveness of the proposed VTransE network by answering the following questions. \textbf{Q1}: Is the idea of embedding relations effective in the visual domain? \textbf{Q2}: What are the effects of the features in relation detection and knowledge transfer? \textbf{Q3}: How does the overall VTransE network perform compared to the other state-of-the-art visual relation models? 

\subsection{Datasets and Metrics}
To the best of our knowledge, there are only two datasets for visual relation detection at a large scale. We used both:
\\
\textbf{VRD}. It is the Visual Relationships dataset~\cite{lu2016visual}. It contains 5,000 images with 100 object categories and 70 predicates. In total, VRD contains 37,993 relation annotations with 6,672 unique relations and 24.25 predicates per object category. We followed the same train/test split as in~\cite{lu2016visual}, \ie, 4,000 training images and 1,000 test images, where 1,877 relationships are only in the test set for zero-shot evaluations.  
\\
\textbf{VG}. It is the latest Visual Genome Version 1.2 relation dataset~\cite{krishna2016visual}. Unlike VRD that is constructed by computer vision experts, VG is annotated by crowd workers and thus the objects and relations are noisy. Therefore, we contact the authors for an official pruning of them. For example, ``young woman'' and ``lady'' are merged to the WordNet hypernym ``woman''. We filtered out relations with less than 5 samples. In summary, VG contains 99,658 images with 200 object categories and 100 predicates, resulting in 1,174,692 relation annotations with 19,237 unique relations and 57 predicates per object category. We split the data into 73,801 for training and 25,857 for testing.

Following~\cite{lu2016visual}, we used Recall@50 (\textbf{R@50}) and Recall@100 (\textbf{R@100}) as evaluation metrics for detection. R@K computes the fraction of times a true relation is predicted in the top K confident relation predictions in an image. Note that precision and average precision (AP) are not proper metrics as visual relations are labeled incompletely and they will penalize the detection if we do not have that particular ground truth. For the relation retrieval task (cf. Section~\ref{sec:q3}), we adopted the Recall rate@5 (\textbf{Rr@5}), which computes the fraction of times the correct result was found among the top 5, and Median rank (\textbf{Med r}), which is the median rank of the first correctly retrieved image~\cite{johnson2015image}. In fact, for datasets with more complete annotations (\eg, VG), even if the recall is low, the actual precision could be high since the number of ground truth in an image is usually larger than 50/100. Therefore, the retrieval task measured by Rr@5 and Med r provides a complementary evaluation.

\subsection{Evaluations of Translation Embedding (Q1)}
\begin{table}[t]
	\centering
	\caption{Predicate prediction performances of the two methods.}
	\label{tab:1}
		\scalebox{0.7}{
	\begin{tabular}{|c|c|c||c|c|}
		\hline
		Method     & \multicolumn{2}{c||}{JointBox} & \multicolumn{2}{c|}{VTransE} \\ \hline
		Dataset    & VRD              & VG            & VRD            & VG            \\ \hline
		R@50  & 25.78            & 46.59             & \textbf{44.76}            & \textbf{62.63}           \\ \hline
		R@100 & 25.78            & 46.77          & \textbf{44.76}            & \textbf{62.87}           \\ \hline
	\end{tabular}}
\end{table}

\begin{figure}
	\centering
	\includegraphics[width=.7\linewidth]{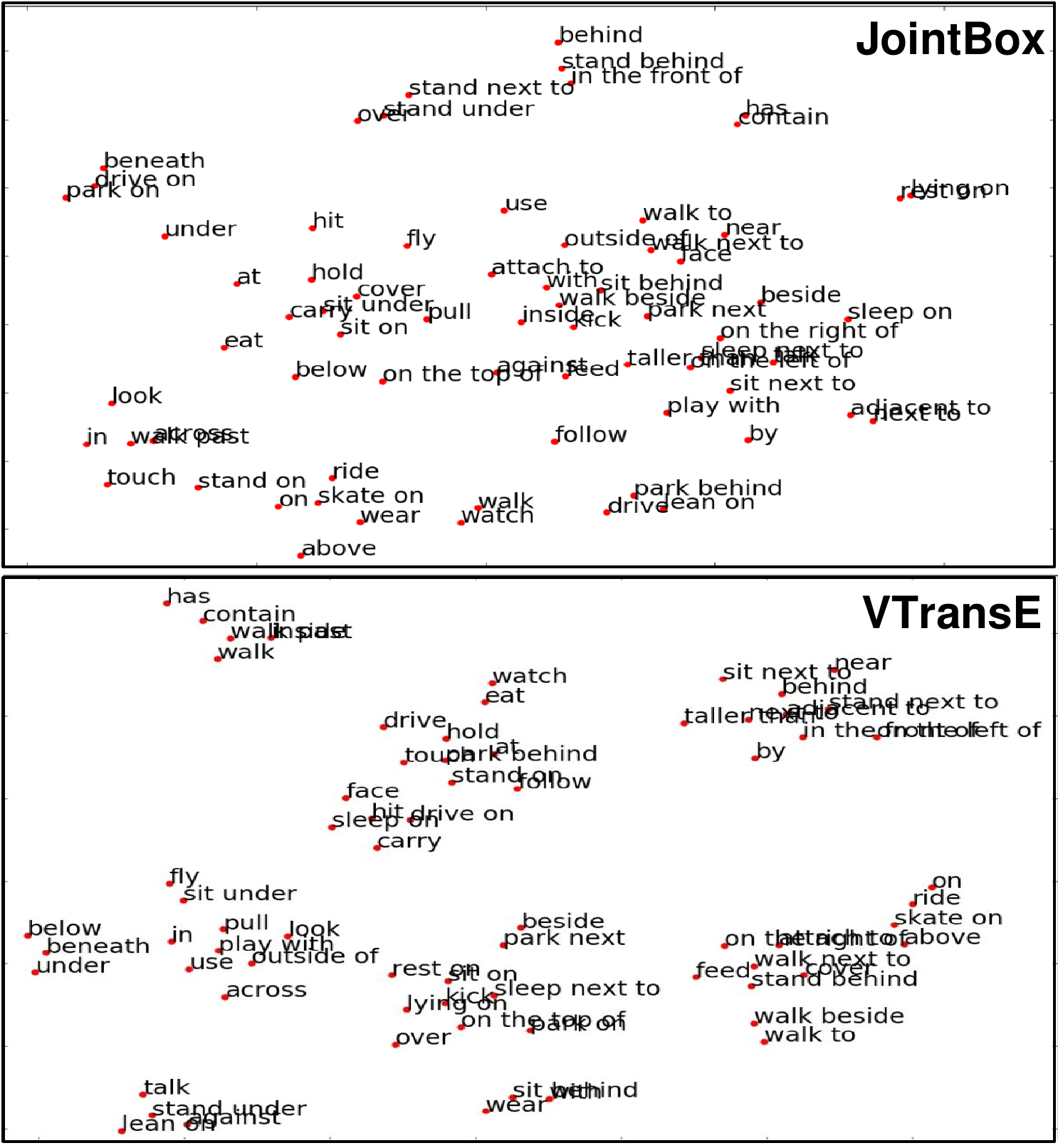}
	\caption{t-SNE visualizations~\cite{maaten2008visualizing} of the 70 predicate model parameters of JointBox and VTransE from VRD. Please zoom in.}
	\vspace{-4mm}
	\label{fig:5}
\end{figure}
\textbf{Setup}.
Visual relation detection requires both object detection and predicate prediction. To investigate whether VTransE is a good model for relations, we need to isolate it from object detection and perform the task of \textbf{Predicate Prediction}: predicting predicates given the ground-truth objects with bounding boxes.

\textbf{Comparing Methods}.
We compared 1) \textbf{JointBox}, a softmax classifier that classifies the images of the \texttt{subject} and \texttt{object} joint bounding boxes into predicates, and 2) \textbf{VTransE} that classifies the predicate of a pair of \texttt{subject} and \texttt{object} boxes. For fair comparison, we only use the RoI pooling visual features of boxes for the two methods. Note that JointBox represents many visual relation models in predicate prediction~\cite{farhadi2010every,lu2016visual,ramanathan2015learning,sadeghi2011recognition} 
 
\textbf{Results}. From Table~\ref{tab:1}, we can see that VTransE formulated in Eq.~\eqref{eq:3} outperforms conventional visual models like JointBox. This is because the predicate model parameters of VTransE---the translation vectors---are able to capture the essential meanings of relations between two objects mapped into a low-dimensional relation space. Figure~\ref{fig:4} illustrates that VTransE can predict correct predicates with diversity while JointBox is more likely to bias on certain visual patterns. For example, JointBox limits \texttt{park on} in cars, but VTransE can generalize to other subjects like plane and bus. Moreover, by inspecting the semantic affinities between the predicate parameter vectors in Figure~\ref{fig:5}, we can speculate that JointBox does not actually model relations but the joint object co-occurrence. For example, in JointBox, the reason why \texttt{beneath} is close to \texttt{drive on} and \texttt{park on} is largely due to the co-occurrence of \texttt{road}-\texttt{beneath}-\texttt{car} and \texttt{car}-\texttt{drive on}-\texttt{road}; however, VTransE is more likely to understand the meaning of \texttt{beneath} as its neighbors are \texttt{below} and \texttt{under}, and it is far from \texttt{on} and \texttt{above}.

\subsection{Evaluations of Features (Q2)}
\textbf{Setup}.
We evaluated how the features proposed in Section~\ref{sec:feat} affect visual relation detection. We performed \textbf{Relation Detection}~\cite{lu2016visual,sadeghi2011recognition}: the input is an image and the output is a set of relation triplets and localizations of both \texttt{subject} and \texttt{object} in the image having at least 0.5 overlap with their ground-truth boxes simultaneously.

\textbf{Comparing Methods}.
We ablated VTransE into four methods in terms of using different features: 1) \textbf{Classeme}, 2) \textbf{Location}, 3) \textbf{Visual}, and 4) \textbf{All} that uses classeme, locations, visual features, and the fusion of the above with a scaling layer (cf. Figure~\ref{fig:3}), respectively. Note that all the above models are trained end-to-end including the object detection module. To further investigate the feature influence on relations, we categorized the predicates into four categories: verb, spatial, preposition and comparative (cf. Supplementary Material for the detailed category list).

\begin{figure}
	\centering
	\includegraphics[width=1\linewidth]{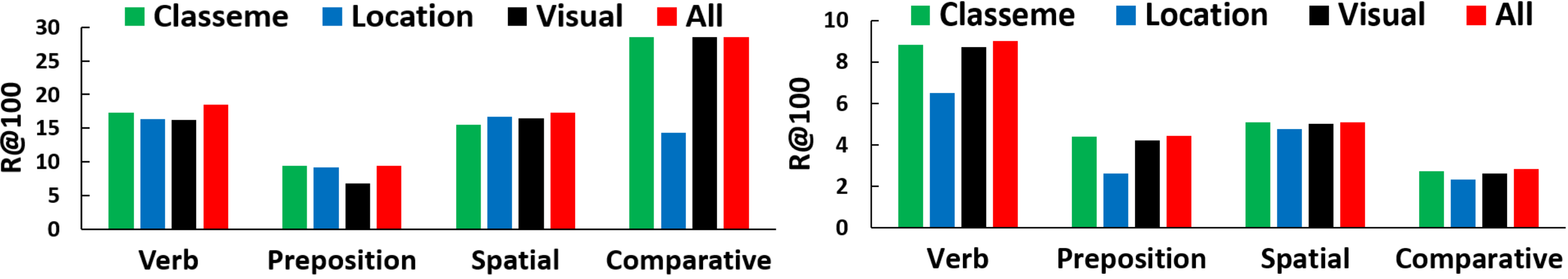}
	\caption{Performances (R@100\%) of relation detection of the four relation types using the four ablated VTransE methods from VRD (left) and VG (right).}
	\vspace{-4mm}
	\label{fig:6}
\end{figure}

\textbf{Results}.
From Figure~\ref{fig:6}, we can see the details of what features are good at detecting what relations: 1) fusing all the features with a learned scaling layer can achieve the best performance on all types of relations; 2) classeme can generally outperform visual features in various kinds of relations as it characterizes both the high-level visual appearances (\eg, what an object looks like) and composition priors (\eg, \texttt{person} is more likely to \texttt{ride}-\texttt{bike} than \texttt{cat}); 3) for spatial relations, location features are better; however, for preposition relations, all features perform relatively poor. This is because the spatial and visual cues of prepositions are volatile such as \texttt{person}-\texttt{with}-\texttt{watch} and \texttt{car}-\texttt{with}-\texttt{wheel}.

\begin{table}[t]
	\centering
	\caption{Object detection mAP\% before (Faster-RCNN) and after training VTransE from VRD (100 objects) and VG (200 objects). Low mAP is mainly due to the incomplete object annotation.}
	\label{tab:2}
	\scalebox{.7}{
		\begin{tabular}{|c|c|c|c|}
			\hline
			\multicolumn{2}{|c|}{VRD} & \multicolumn{2}{c|}{VG} \\ \hline
			Before      & After      & Before      & After     \\ \hline
			13.32          & \textbf{13.98}         & 6.21          & \textbf{6.58}        \\ \hline
		\end{tabular}}
	\vspace{-3mm}
	\end{table}

\begin{figure}
	\centering
	\includegraphics[width=.8\linewidth]{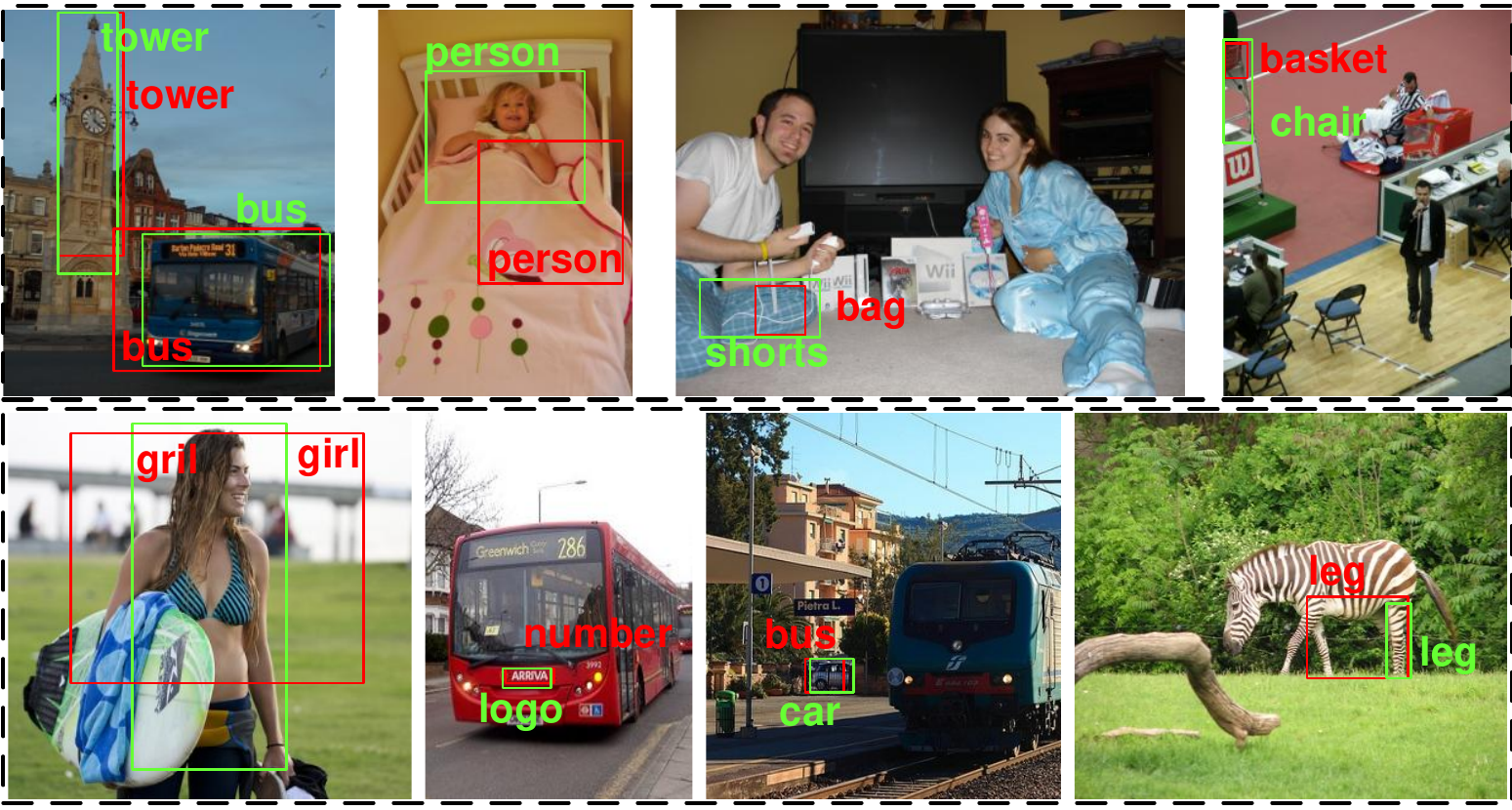}
	\caption{Qualitative object detection examples before (red box and font) and after (green box and font) training VTransE from VRD (top row) and VG (bottom row).}
	\vspace{-6mm}
	\label{fig:7}
\end{figure}

Table~\ref{tab:2} shows that the end-to-end training of VTransE can improve the object detection. This is mainly due to that the proposed feature extraction layer allows knowledge transfer so that the errors made by relation prediction can be back-propagated to the front object detection module. In fact, the improvement can be expected since we incorporate additional relation labels besides object labels. As shown in Figure~\ref{fig:7}, compared to the pre-trained Faster-RCNN module, the object detection module trained by VTransE can generally improve bounding boxes, such as minor refinement or even recovery from drastic dislocation and corrections for wrong detections. This demonstrates that relations place objects in a contextual scene. For example, relation can recover \texttt{shorts} from the wrong detection \texttt{bag}, even though the correct detection should be \texttt{pants}, which is semantically similar to \texttt{shorts}. This correction is likely inferred by the relation \texttt{person}-\texttt{wear}-\texttt{shorts}/\texttt{pants}.

\begin{table*}[t]
	\centering
	\caption{Performances of phrase detection, relation detection, relation retrieval using various methods on both datasets. ``--'' denotes that the result is not applicable. (cf. Supplementary Material for the incomplete annotation in VRD that causes low retrieval performances.)}
	\label{tab:3}
	\scalebox{0.7}{
		\begin{tabular}{|c||c|c|c|c|c|c||c|c|c|c|c|c|}
			\hline
			Dataset      & \multicolumn{6}{c||}{VRD~\cite{lu2016visual}}                                                                      & \multicolumn{6}{c|}{VG~\cite{krishna2016visual}}                                                                      \\ \hline
			Task         & \multicolumn{2}{c|}{Phrase Det.} & \multicolumn{2}{c|}{Relation Det.} & \multicolumn{2}{c||}{Retrieval} & \multicolumn{2}{c|}{Phrase Det.} & \multicolumn{2}{c|}{Relation Det.} & \multicolumn{2}{c|}{Retrieval} \\ \hline
			Metric       & R@50          & R@100         & R@50           & R@100          & Rr@5           & Med r            & R@50           & R@100      & R@50           & R@100       & Rr@5           & Med r        \\ \hline
			VisualPhrase~\cite{sadeghi2011recognition} & 0.54           & 0.63           & --            & --            & 3.51            & 204             & 3.41           & 4.27           & --            & --            & 11.42            & 18             \\ \hline
			DenseCap~\cite{johnson2015densecap}     & 0.62           &  0.77       &--           &--            & 4.16            & 199            & 3.85           & 5.01           &--            &--            & 12.95            & 13             \\ \hline
			Lu's-V~\cite{lu2016visual}       &2.24           &2.61            &1.58            &1.85            &2.82            & 211            & --           & --           & --            & --            & --            & --             \\
			\hline
			Lu's-VLK~\cite{lu2016visual}       & 16.17           & 17.03           & 13.86            & 14.70            & \textbf{8.75}            & 137             & --           & --           & --            & --            & --            & --             \\
			 \hline
			VTransE      & \textbf{19.42}           & \textbf{22.42}           & \textbf{14.07}            & \textbf{15.20}            & 7.89           & \textbf{41}             & \textbf{9.46}           &   \textbf{10.45}         & \textbf{5.52}            &   \textbf{6.04}          & \textbf{14.65}            & \textbf{7}             \\ \hline\hline
			VTransE-2stage & 18.45           & 21.29           & 13.30            & 14.64            & 7.14            & 41             & 8.73           &  10.11           & 4.97            & 5.48            & 12.82            & 12 \\\hline
			\hline
			Random & 0.06           & 0.11           & 7.14$\times$$10^{-3}$            & 1.43$\times$$10^{-2}$            & 2.95            & 497             & 0.04           & 0.07           & 1.25$\times$$10^{-3}$                & 2.50$\times$$10^{-3}$           & 3.45            & 1.28$\times 10^4$ \\ \hline
		\end{tabular}}
	\end{table*}

\begin{figure*}
		\centering
		\includegraphics[width=.9\linewidth]{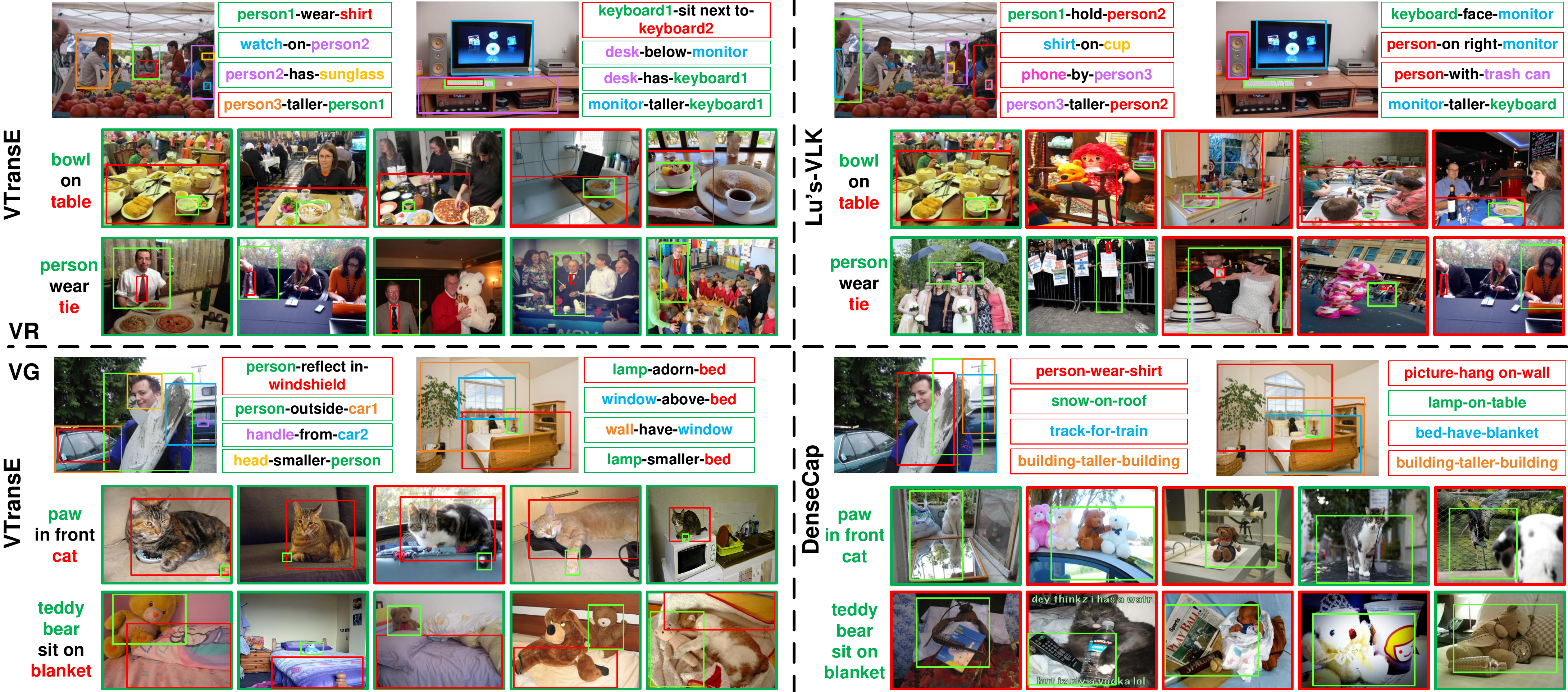}
		\caption{Qualitative examples of relation detection (4 top-1 detections from 4 predicate types) and retrieval (top-5 images). We compare our VTransE with its best competitors: Lu's-VLK on VRD and DenseCap on VG. Green and red borders denote correct and incorrect results.}
		\vspace{-4mm}
		\label{fig:8}
\end{figure*}

\begin{figure*}
	\centering
	\includegraphics[width=.9\linewidth]{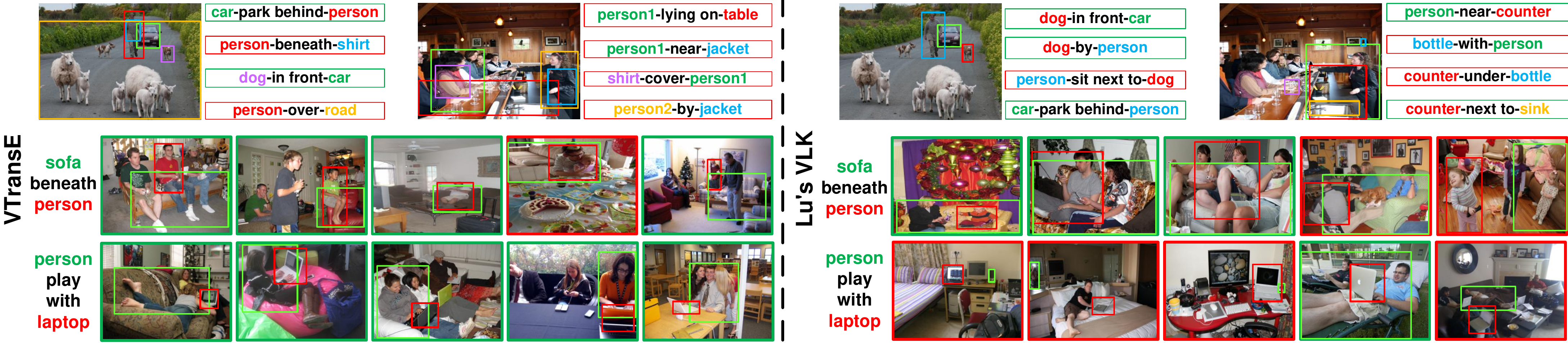}
	\caption{Qualitative examples of zero-shot relation detection (top-4) and retrieval (top-5) using VTransE and Lu's-VLK on VRD. Green and red borders denote correct and incorrect results.}
	\vspace{-3mm}
	\label{fig:9}
\end{figure*}

\subsection{Comparison with State-of-The-Arts (Q3)}\label{sec:q3}
\textbf{Setup}.
As we will introduce later, some joint relation models can only detect a joint bounding box for an entire relation; thus, besides relation detection, we performed \textbf{Phrase Detection}~\cite{lu2016visual}: the input is an image and the output is a set of relation triplets and localizations of the entire bounding box for each relation that having at least 0.5 overlap with the ground-truth joint \texttt{subject} and \texttt{object} box.

For more extensive evaluations, we also performed two additional tasks. 1) \textbf{Relation Retrieval}: image search with the query of a relation triplet. We first detect the relation query in gallery (\ie, test) images and then score them according to the average detection scores of the query relation. An image with at least one successful query relation detection is considered as a hit. This task is a representation of the compositional semantic retrieval~\cite{johnson2015image}; We selected the top 1,000 frequent relations as queries. 2) \textbf{Zero-shot Learning}~\cite{lu2016visual}: individual \texttt{subject}, \texttt{object}, and \texttt{predicate} are seen in both training and test, but some specific triplet compositions are only in the test set. Due to the long-tailed relation distribution, it is a practical setting since it is impossible to collect data for all triplets. 

\textbf{Comparing Methods}.
We compared the VTransE network to four state-of-the-art visual relation detection models. 1) \textbf{VisualPhrase}~\cite{sadeghi2011recognition}: a joint relation model that considers every unique relation triplet as an relation class. For fair comparison, we replace the original DPM object detection model~\cite{felzenszwalb2010object} with Faster-RCNN~\cite{ren2015faster}; 2) \textbf{DenseCap}~\cite{johnson2015densecap}: it detects sub-image regions and generate their descriptions simultaneously. It is an end-to-end model using bilinear interpolated visual features for region localizations. We replace its LSTM classification layer with softmax for relation prediction. Thus, it can be considered as an joint relation model; 3) \textbf{Lu's-V} (V-only in~\cite{lu2016visual}): it is a two-stage separate model that first uses R-CNN~\cite{girshick2014rich} for object detection and then adopts a large-margin JointBox model for predicate classification; 4) \textbf{Lu's-VLK} (V+L+K in~\cite{lu2016visual}): a two-stage separate model that combines Lu's-V and word2vec language priors~\cite{mikolov2013distributed}. In addition, we compared VTransE to its two-stage training model \textbf{VTransE-2stage} that apply Faster-RCNN for object detection and then perform predicate predication using translation embedding as in Q1.

As we have no training source codes of Lu's methods, we cannot apply them in VG and we quoted the results of VRD reported in their paper~\cite{lu2016visual}. Moreover, as the joint relation models such as VisualPhrase and DenseCap can only detect relation triplet as a whole, they are not applicable in zero-shot learning. Therefore, we only report zero-shot results (detection and retrieval) on VRD for the official 1,877 zero-shot relations~\cite{lu2016visual}. 

\textbf{Results}. From the quantitative results in Table~\ref{tab:3} and the qualitative results in Figure~\ref{fig:8}, we have:
\\
1) Separate relation models like VTransE and Lu's-V outperform joint models like VisualPhrase and DenseCap significantly, especially on VRD. This is because the classification space of joint models for all possible relationships is large (\eg, 6,672 and 19,237 training relations in VRD 
and VG, respectively), leading to insufficient samples for training infrequent relations. 
\\
2) For separate models, better object detection networks, such as Faster-RCNN v.s. R-CNN used in VTrasnE and Lu's, are beneficial for relation detections. As shown in Figure~\ref{fig:8}, on VRD dataset, Lu's-VLK mistakes \texttt{soundbox} as \texttt{person} and \texttt{plate} as \texttt{bowl}. We believe that this is a significant reason why their visual model Lu's-V is considerably worse than VTransE. 
\\
3) Even though VTransE is a purely visual model, we can still outperform Lu's-VLK which incorporates language priors, \eg, on VRD measured by R@50 and Med r, we are 20\%, 2\%, and 230\% relatively better in phrase detection, relation detection, and relation retrieval, respectively. First, the classeme feature can serve as a similar role as language priors. Second, location feature is indispensable to relations. Take the \texttt{person}-\texttt{wear}-\texttt{tie} relation query as an example in Figure~\ref{fig:8}, when there are multiple \texttt{person} detections in an image, Lu's-VLK usually relates \texttt{tie} to the wrong \texttt{person}, regardless the fact that the spatial distance is far. Similar examples can be also found in the false detection \texttt{shirt}-\texttt{on}-\texttt{cup} of Lu's-VLK. 
\\
4) The end-to-end VTransE is better than VTransE-2stage across all the tasks on both datasets. Together with the results in Q2, they demonstrate the effectiveness of reciprocal learning between objects and relations.

From the zero-shot quantitative results in Table~\ref{tab:4} and the qualitative results in Figure~\ref{fig:9}, we have:
\\
1) The performances of ours and the compared methods degrade drastically, \eg, for relation detection, VTransE and Lu's-VLK suffer 88\% and 79\% performance (R@100) drop, respectively. This is the key \textbf{limitation} of VTransE. Perhaps this is because our transformation from feature space to relation space in Eq.~\eqref{eq:1} is too generic, especially for verbs, and thus fails to capture the relation-specific visual deformations. For example, VTransE cannot discriminate between \texttt{person}-\texttt{lying on}-\texttt{table} and \texttt{person}-\texttt{sit next to}-\texttt{table}. One remedy is to incorporate predicate and object models~\cite{mallya2016learning}, although it will increase the model complexity from $\mathcal{O}(N+R)$ to $\mathcal{O}(NR)$, where $N$ is the number of objects and $R$ is the number of predicates. 
\\
2) Both as visual models, our VTransE is significantly better than Lu's-V in zero-shot relation predictions; nevertheless, as a multi-modal model, Lu's-VLK surpasses VTransE by exploiting language priors. But, since visual relations are volatile to specific examples, language priors are not always correct---Lu's-VLK can be misled by frequent language collocations which are invalid in visual examples, \eg, the mismatch of \texttt{subject} and \texttt{object} in \texttt{sofa}-\texttt{beneath}-\texttt{person} and \texttt{person}-\texttt{play with}-\texttt{laptop}.

\begin{table}[t]
	\centering
	\caption{Performances of zero-shot phrase detection, relation detection, relation retrieval using various methods on VRD. Note that joint models like VisualPhrase and DenseCap do not apply in zero-shot setting.}
	\label{tab:4}
	\scalebox{0.6}{
		\begin{tabular}{|c||c|c||c|c||c|c|}
			\hline
			Task         & \multicolumn{2}{c||}{Phrase Det.} & \multicolumn{2}{c||}{Relation Det.} & \multicolumn{2}{c|}{Retrieval}\\ \hline
			Metric       & R@50          & R@100         & R@50           & R@100          & Rr@5           & Med r \\ \hline
			Lu's-V~\cite{lu2016visual}        & 0.95           &1.12            &0.67             & 0.78            & 0.54            & 454  \\ \hline
			Lu's-VLK~\cite{lu2016visual}      & \textbf{3.36}           & \textbf{3.75}           & \textbf{3.13}            & \textbf{3.52}            & 1.24            & 434   \\ \hline
			VTransE      & 2.65           &3.51         &1.71            &2.14            & \textbf{1.42}            & \textbf{422}   \\ \hline
			\hline
			Random      & 0.02           & 0.04           &  7.14$\times$$10^{-3}$          & 1.43$\times$$10^{-2}$            & 0.45            & 499   \\ \hline
		\end{tabular}}
		\vspace{-4mm}
	\end{table}
\vspace{-2mm}
\section{Conclusions}
\vspace{-2mm}
We focused on the \emph{visual relation detection} task that is believed to offer a comprehensive scene understanding for connecting computer vision and natural language. Towards this task we introduced the VTransE network for simultaneous object detection and relation prediction. VTransE is an end-to-end and fully-convolutional architecture that consists of an object detection module, a novel differentiable feature extraction layer, and a novel visual translation embedding layer for predicate classification. Moving forward, we are going to 1) model higher-order relations such as \texttt{person}-\texttt{throw}-\texttt{ball}-\texttt{to}-\texttt{dog}, 2) tackle the challenge of zero-shot relation learning, and 3) apply VTransE in a VQA system based on relation reasoning. 

\section{Supplementary Material}
\begin{figure*}[!tp]
	\centering
	\includegraphics[width=1\linewidth]{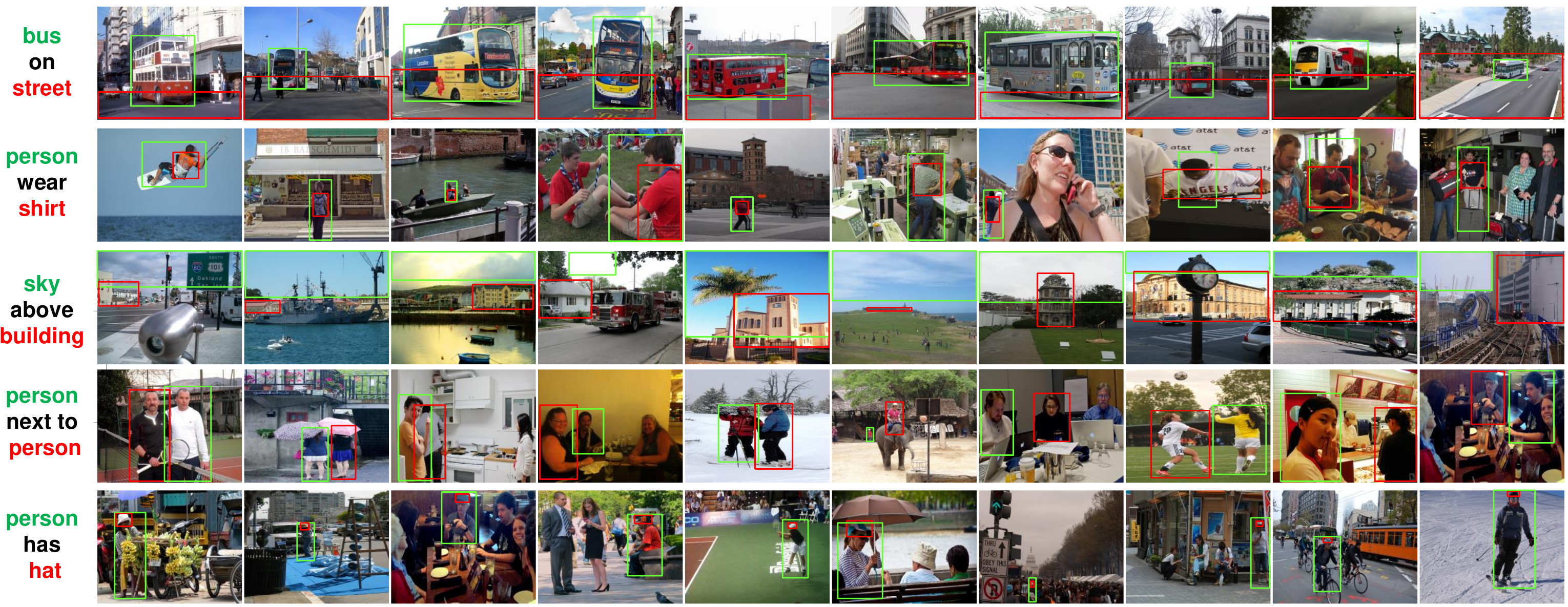}
	\caption{Top 10 VRD images retrieved by the query relations on the left. Note that none of the images is labeled as the corresponding query.}
	\label{fig:1s}
\end{figure*}
\subsection{Predicate Categorization}
\subsubsection{70 predicates in Visual Relationship Dataset}
\textbf{Verb}: attach to, carry, contain, cover, drive, drive on, eat, face, feed, fly, follow, hit, hold, kick, lean on, look, lying on, park behind, park next, park on, play with, pull, rest on, ride, sit behind, sit next to, sit on, sit under, skate on, sleep next to, sleep on, stand behind, stand next to, stand on, stand under, talk, touch, use, walk, walk beside, walk next to, walk past, walk to, watch, wear.

\textbf{Spatial}: above, adjacent to, behind, below, beneath, beside, in, in the front of, inside, near, next to, on, on the left of, on the right of, on the top of, outside of, over, under.

\textbf{Preposition}: across, against, at, by, has, with.

\textbf{Comparative}: taller than.

\subsubsection{100 predicates in Visual Genome Dataset}
\textbf{Verb}: adorn, attach to, belong to, build into, carry, cast, catch, connect to, contain, cover, cover in, cover with, cross, cut, drive on, eat, face, fill with, fly, fly in, grow in, grow on, hang in, hang on, hit, hold, hold by, lay in, lay on, lean on, look at, mount on, paint on, park, play, print on, pull, read, reflect in, rest on, ride, say, show, sit at, sit in, sit on, stand behind, stand on, standing by, standing in, standing next to, support, surround, swing, throw, touch, use, walk, walk in, walk on, watch, wear, wear by, write on.

\textbf{Spatial}: above, behind, below, beneath, between, in, in front of, in middle of, inside, near, next to, on, on back of, on bottom of, on side of, on top of, outside, over, under, underneath.

\textbf{Preposition}: across, against, along, around, at, beside, by, for, from, have, of, part of, to, with.

\textbf{Comparative}: small than, tall than.
\begin{figure*}[h]
	\centering
	\includegraphics[width=.8\linewidth]{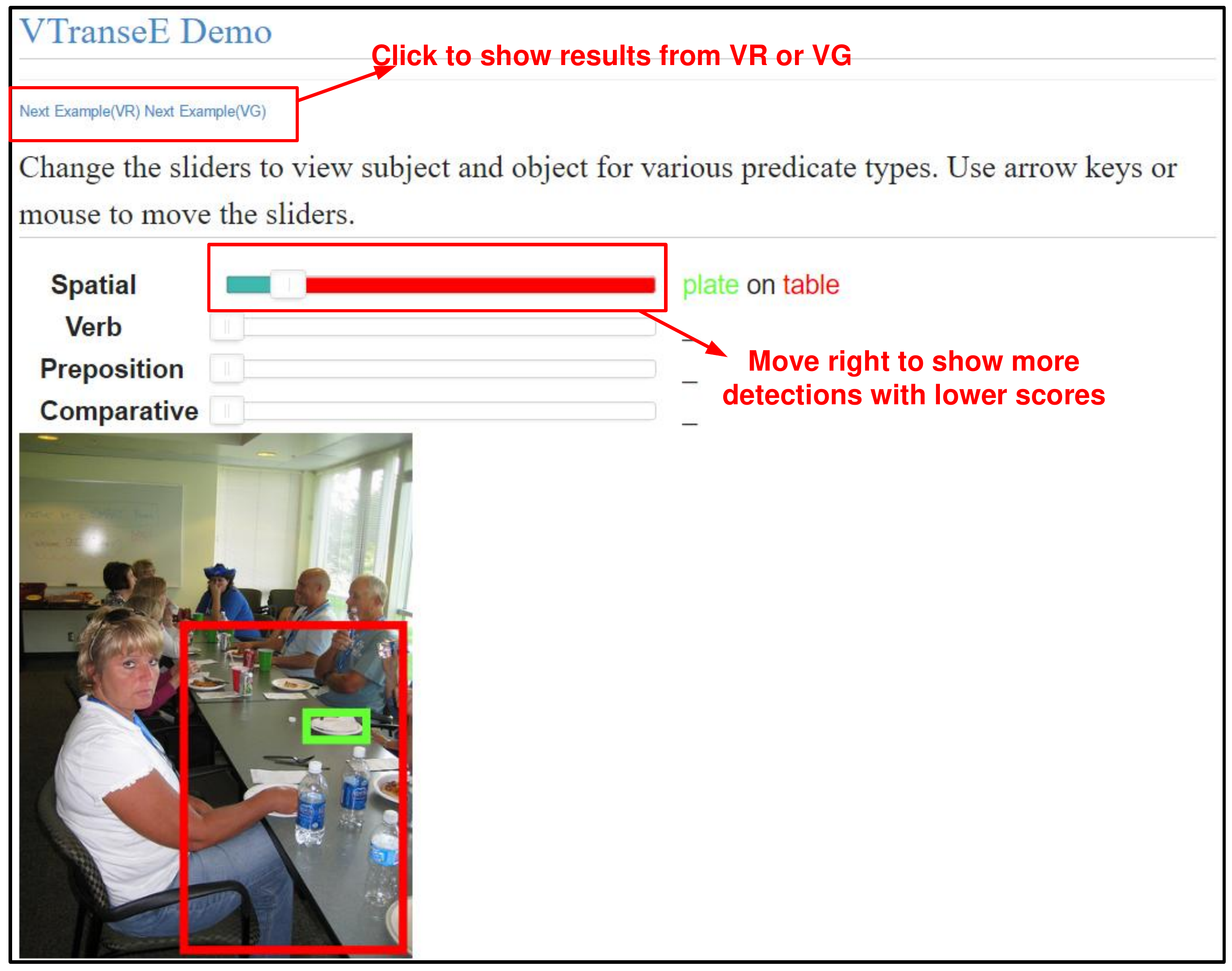}
	\caption{The interface of the VTransE demo.}
	\label{fig:2s}
\end{figure*}
\subsection{Incomplete Annotation in VRD}
More qualitative relation retrieval results from VRD are shown in Figure~\ref{fig:1s}. We can see that even though most of the top 10 results are correct, none of them is labeled as the corresponding query relation. We believe that this incomplete annotation in VRD is the main cause of the low retrieval performance, as reported in Table 3 and Table 4 from the paper. 

\subsection{Demo Link}
We provide a demo running on an Amazon server (US East N. Virginia)
) \url{cvpr.zl.io}. The interface is illustrated in Figure~\ref{fig:2s}. Note that for clear visualizations, only detections scored about top 20\% of each predicate type are shown.

{\footnotesize
\bibliographystyle{ieee}
\bibliography{egbib}

\begin{thebibliography}{10}\itemsep=-1pt

\bibitem{akata2015label}
Z.~Akata, F.~Perronnin, Z.~Harchaoui, and C.~Schmid.
\newblock Label-embedding for image classification.
\newblock {\em TPAMI}, 2015.

\bibitem{andreas2016deep}
J.~Andreas, M.~Rohrbach, T.~Darrell, and D.~Klein.
\newblock Deep compositional question answering with neural module networks.
\newblock In {\em CVPR}, 2016.

\bibitem{antol2015vqa}
S.~Antol, A.~Agrawal, J.~Lu, M.~Mitchell, D.~Batra, C.~Lawrence~Zitnick, and
  D.~Parikh.
\newblock Vqa: Visual question answering.
\newblock In {\em ICCV}, 2015.

\bibitem{atzmon2016learning}
Y.~Atzmon, J.~Berant, V.~Kezami, A.~Globerson, and G.~Chechik.
\newblock Learning to generalize to new compositions in image understanding.
\newblock In {\em EMNLP}, 2016.

\bibitem{bernardi2016automatic}
R.~Bernardi, R.~Cakici, D.~Elliott, A.~Erdem, E.~Erdem, N.~Ikizler-Cinbis,
  F.~Keller, A.~Muscat, and B.~Plank.
\newblock Automatic description generation from images: A survey of models,
  datasets, and evaluation measures.
\newblock {\em JAIR}, 2016.

\bibitem{bordes2013translating}
A.~Bordes, N.~Usunier, A.~Garcia-Duran, J.~Weston, and O.~Yakhnenko.
\newblock Translating embeddings for modeling multi-relational data.
\newblock In {\em NIPS}, 2013.

\bibitem{chao2000representation}
L.~L. Chao and A.~Martin.
\newblock Representation of manipulable man-made objects in the dorsal stream.
\newblock {\em Neuroimage}, 2000.

\bibitem{desai2011discriminative}
C.~Desai, D.~Ramanan, and C.~C. Fowlkes.
\newblock Discriminative models for multi-class object layout.
\newblock {\em IJCV}, 2011.

\bibitem{dong2015question}
L.~Dong, F.~Wei, M.~Zhou, and K.~Xu.
\newblock Question answering over freebase with multi-column convolutional
  neural networks.
\newblock In {\em ACL}, 2015.

\bibitem{farhadi2010every}
A.~Farhadi, M.~Hejrati, M.~A. Sadeghi, P.~Young, C.~Rashtchian, J.~Hockenmaier,
  and D.~Forsyth.
\newblock Every picture tells a story: Generating sentences from images.
\newblock In {\em ECCV}, 2010.

\bibitem{felzenszwalb2010object}
P.~F. Felzenszwalb, R.~B. Girshick, D.~McAllester, and D.~Ramanan.
\newblock Object detection with discriminatively trained part-based models.
\newblock {\em TPAMI}, 2010.

\bibitem{frome2013devise}
A.~Frome, G.~S. Corrado, J.~Shlens, S.~Bengio, J.~Dean, T.~Mikolov, et~al.
\newblock Devise: A deep visual-semantic embedding model.
\newblock In {\em NIPS}, 2013.

\bibitem{girshick2015fast}
R.~Girshick.
\newblock Fast r-cnn.
\newblock In {\em ICCV}, 2015.

\bibitem{girshick2014rich}
R.~Girshick, J.~Donahue, T.~Darrell, and J.~Malik.
\newblock Rich feature hierarchies for accurate object detection and semantic
  segmentation.
\newblock In {\em CVPR}, 2014.

\bibitem{gregor2015draw}
K.~Gregor, I.~Danihelka, A.~Graves, D.~J. Rezende, and D.~Wierstra.
\newblock Draw: A recurrent neural network for image generation.
\newblock {\em arXiv preprint arXiv:1502.04623}, 2015.

\bibitem{gupta2008beyond}
A.~Gupta and L.~S. Davis.
\newblock Beyond nouns: Exploiting prepositions and comparative adjectives for
  learning visual classifiers.
\newblock In {\em ECCV}, 2008.

\bibitem{gupta2009observing}
A.~Gupta, A.~Kembhavi, and L.~S. Davis.
\newblock Observing human-object interactions: Using spatial and functional
  compatibility for recognition.
\newblock {\em TPAMI}, 2009.

\bibitem{he2015deep}
K.~He, X.~Zhang, S.~Ren, and J.~Sun.
\newblock Deep residual learning for image recognition.
\newblock 2016.

\bibitem{jabri2016revisiting}
A.~Jabri, A.~Joulin, and L.~van~der Maaten.
\newblock Revisiting visual question answering baselines.
\newblock In {\em ECCV}, 2016.

\bibitem{jaderberg2015spatial}
M.~Jaderberg, K.~Simonyan, A.~Zisserman, et~al.
\newblock Spatial transformer networks.
\newblock In {\em NIPS}, 2015.

\bibitem{johnson2015densecap}
J.~Johnson, A.~Karpathy, and L.~Fei-Fei.
\newblock Densecap: Fully convolutional localization networks for dense
  captioning.
\newblock In {\em CVPR}, 2016.

\bibitem{johnson2015image}
J.~Johnson, R.~Krishna, M.~Stark, L.-J. Li, D.~A. Shamma, M.~S. Bernstein, and
  L.~Fei-Fei.
\newblock Image retrieval using scene graphs.
\newblock In {\em CVPR}, 2015.

\bibitem{karpathy2015deep}
A.~Karpathy and L.~Fei-Fei.
\newblock Deep visual-semantic alignments for generating image descriptions.
\newblock In {\em CVPR}, 2015.

\bibitem{kingma2014adam}
D.~Kingma and J.~Ba.
\newblock Adam: A method for stochastic optimization.
\newblock {\em arXiv preprint arXiv:1412.6980}, 2014.

\bibitem{krishna2016visual}
R.~Krishna, Y.~Zhu, O.~Groth, J.~Johnson, K.~Hata, J.~Kravitz, S.~Chen,
  Y.~Kalantidis, L.-J. Li, D.~A. Shamma, et~al.
\newblock Visual genome: Connecting language and vision using crowdsourced
  dense image annotations.
\newblock {\em IJCV}, 2016.

\bibitem{lin2015learning}
Y.~Lin, Z.~Liu, M.~Sun, Y.~Liu, and X.~Zhu.
\newblock Learning entity and relation embeddings for knowledge graph
  completion.
\newblock In {\em AAAI}, 2015.

\bibitem{liu2015ssd}
W.~Liu, D.~Anguelov, D.~Erhan, C.~Szegedy, and S.~Reed.
\newblock Ssd: Single shot multibox detector.
\newblock In {\em ECCV}, 2016.

\bibitem{lu2016visual}
C.~Lu, R.~Krishna, M.~Bernstein, and L.~Fei-Fei.
\newblock Visual relationship detection with language priors.
\newblock In {\em ECCV}, 2016.

\bibitem{maaten2008visualizing}
L.~v.~d. Maaten and G.~Hinton.
\newblock Visualizing data using t-sne.
\newblock {\em JMLR}, 2008.

\bibitem{mallya2016learning}
A.~Mallya and S.~Lazebnik.
\newblock Learning models for actions and person-object interactions with
  transfer to question answering.
\newblock In {\em ECCV}, 2016.

\bibitem{mikolov2013distributed}
T.~Mikolov, I.~Sutskever, K.~Chen, G.~S. Corrado, and J.~Dean.
\newblock Distributed representations of words and phrases and their
  compositionality.
\newblock In {\em NIPS}, 2013.

\bibitem{nickel2016review}
M.~Nickel, K.~Murphy, V.~Tresp, and E.~Gabrilovich.
\newblock A review of relational machine learning for knowledge graphs.
\newblock {\em Proceedings of the IEEE}, 2016.

\bibitem{plummer2015flickr30k}
B.~A. Plummer, L.~Wang, C.~M. Cervantes, J.~C. Caicedo, J.~Hockenmaier, and
  S.~Lazebnik.
\newblock Flickr30k entities: Collecting region-to-phrase correspondences for
  richer image-to-sentence models.
\newblock In {\em ICCV}, 2015.

\bibitem{ramanathan2015learning}
V.~Ramanathan, C.~Li, J.~Deng, W.~Han, Z.~Li, K.~Gu, Y.~Song, S.~Bengio,
  C.~Rossenberg, and L.~Fei-Fei.
\newblock Learning semantic relationships for better action retrieval in
  images.
\newblock In {\em CVPR}, 2015.

\bibitem{redmon2015you}
J.~Redmon, S.~Divvala, R.~Girshick, and A.~Farhadi.
\newblock You only look once: Unified, real-time object detection.
\newblock In {\em CVPR}, 2016.

\bibitem{ren2015faster}
S.~Ren, K.~He, R.~Girshick, and J.~Sun.
\newblock Faster r-cnn: Towards real-time object detection with region proposal
  networks.
\newblock In {\em NIPS}, 2015.

\bibitem{sadeghi2015viske}
F.~Sadeghi, S.~K. Divvala, and A.~Farhadi.
\newblock Viske: Visual knowledge extraction and question answering by visual
  verification of relation phrases.
\newblock In {\em CVPR}, 2015.

\bibitem{sadeghi2011recognition}
M.~A. Sadeghi and A.~Farhadi.
\newblock Recognition using visual phrases.
\newblock In {\em CVPR}, 2011.

\bibitem{simonyan2014very}
K.~Simonyan and A.~Zisserman.
\newblock Very deep convolutional networks for large-scale image recognition.
\newblock {\em arXiv preprint arXiv:1409.1556}, 2014.

\bibitem{torresani2010efficient}
L.~Torresani, M.~Szummer, and A.~Fitzgibbon.
\newblock Efficient object category recognition using classemes.
\newblock In {\em ECCV}, 2010.

\bibitem{vinyals2016show}
O.~Vinyals, A.~Toshev, S.~Bengio, and D.~Erhan.
\newblock Show and tell: Lessons learned from the 2015 mscoco image captioning
  challenge.
\newblock {\em TPAMI}, 2016.

\bibitem{wang2016fvqa}
P.~Wang, Q.~Wu, C.~Shen, A.~v.~d. Hengel, and A.~Dick.
\newblock Fvqa: Fact-based visual question answering.
\newblock {\em arXiv preprint arXiv:1606.05433}, 2016.

\bibitem{yao2010modeling}
B.~Yao and L.~Fei-Fei.
\newblock Modeling mutual context of object and human pose in human-object
  interaction activities.
\newblock In {\em CVPR}, 2010.

\end{thebibliography}
}

\end{document}